\documentclass[final]{cvpr}

\usepackage{times}
\usepackage{epsfig}
\usepackage{graphicx}
\usepackage{amsmath}
\usepackage{amssymb}
\usepackage{multirow}

\usepackage[pagebackref=true,breaklinks=true,colorlinks,bookmarks=false]{hyperref}

\graphicspath{{figs/}{evaluation/}}

\renewcommand{\vec}[1]{\mathbf{#1}}

\def\cvprPaperID{****} % *** Enter the CVPR Paper ID here
\def\confYear{CVPR 2021}

\begin{document}

%%%%%%%%% TITLE
\title{Embedded Vision for Self-Driving on Forest Roads}

\author{Sorin Grigorescu, Mihai Zaha, Bogdan Trasnea and Cosmin Ginerica\\
\\
RovisLab (Robotics, Vision and Control Laboratory), \url{www.rovislab.com}\\
Mihai Viteazu 5, 500174 Brasov, Romania\\
{\tt\small contact@rovislab.com}
%\and
}

\maketitle

%%%%%%%%% ABSTRACT
\begin{abstract}
   Forest roads in Romania are unique natural wildlife sites used for recreation by countless tourists. In order to protect and maintain these roads, we propose RovisLab AMTU (Autonomous Mobile Test Unit), which is a robotic system designed to autonomously navigate off-road terrain and inspect if any deforestation or damage occurred along tracked route. AMTU's core component is its embedded vision module, optimized for real-time environment perception. For achieving a high computation speed, we use a learning system to train a multi-task Deep Neural Network (DNN) for scene and instance segmentation of objects, while the keypoints required for simultaneous localization and mapping are calculated using a handcrafted FAST feature detector and the Lucas-Kanade tracking algorithm. Both the DNN and the handcrafted backbone are run in parallel on the GPU of an NVIDIA AGX Xavier board. We show experimental results on the test track of our research facility. Multimedia material is available at:
   
   \url{https://youtu.be/PKl30NtzAWE}
   
   \url{https://youtu.be/dXxlpvzDsSw}
\end{abstract}

\vspace{-2em}

%%%%%%%%% BODY TEXT
\section{Introduction}
\label{sec:introduction}

\begin{figure*}
	\centering
	\begin{center}
		\includegraphics[scale=0.9]{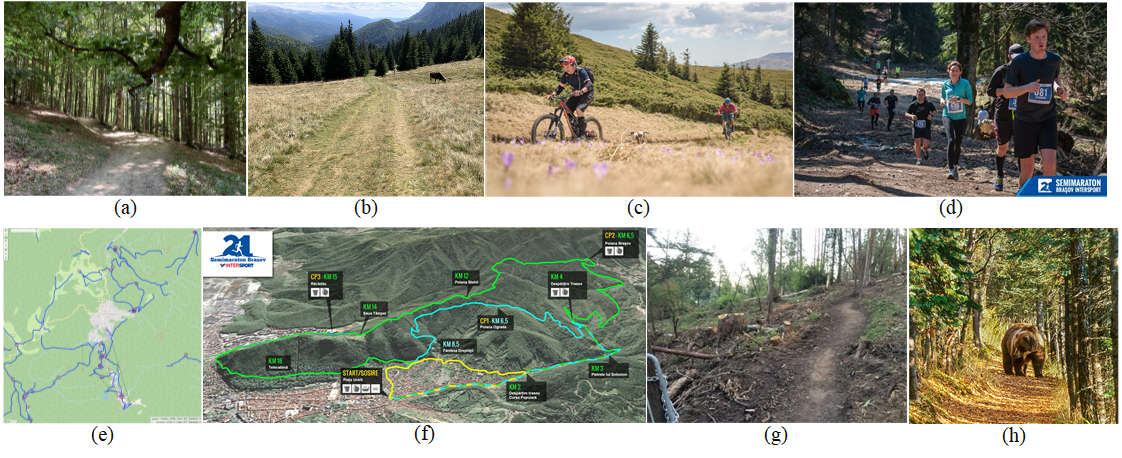}
		\caption{\textbf{Snapshots and maps of forest roads in Romania.} (a,b) Poiana Izvoarelor, Busteni, Romania. (c) Biking around Viscri village in Transylvania, Romania. (d) Half-marathon trail contest on a forest road in Brasov, Romania. (e) Density of forest roads around Brasov area, Romania. (f) Map of a marathon trail running contest in Brasov, Romania, taking place largely on forest roads. (g) Deforestation near a trail in Piatra Craiului mountain. (h) Bear spotted on a forest road near Brasov, Romania.}
        \label{fig:forestry_roads}
	\end{center}
    \vspace{-1em}
\end{figure*}

With over 25.000 km of forest roads and home of more than $60\%$ of Europe's original forest\footnotetext{\url{https://www.saveparadiseforests.eu/en/primary-forest-in-europe/}}, Romania holds a unique natural wildlife heritage regularly visited by tourists and locals aside. Due to their beauty and high quantity of oxygen, forest roads within or near different cities in Romania are used for recreational activities, such as hiking, cycling, or trail running competitions. A couple of snapshots of from forest roads in Romania are shown in Figure~\ref{fig:forestry_roads}.

\begin{figure}
	\centering
	\begin{center}
		\includegraphics[scale=0.75]{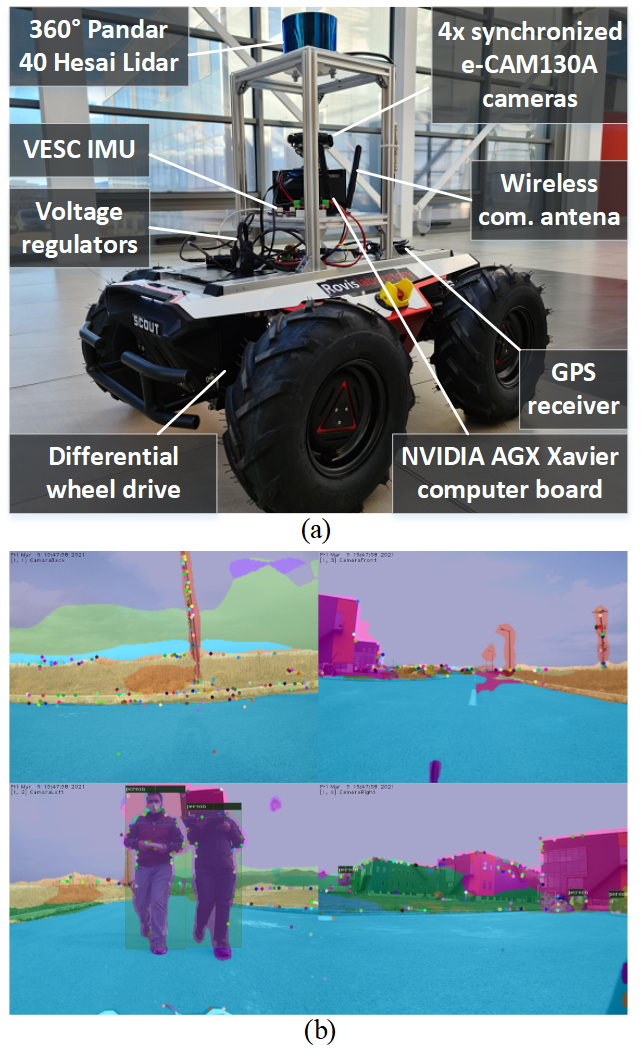}
		\caption{\textbf{RovisLab AMTU (Autonomous Mobile Test Unit)}. (a) Hardware components. (b) Embedded vision results using the four cameras of the robot. The perception system is trained to segment the scene, as well as to track relevant feature points used for simultaneous localization and mapping. The color of the points encodes each feature point's ID (best viewed in color).}
        \label{fig:rovislab_amtu}
	\end{center}
    \vspace{-2em}
\end{figure}

These roads and the forests around them are a unique natural heritage for Europe and Romania, that must be preserved and maintained for the generations to come. Possible threats include deforestation (Figure~\ref{fig:forestry_roads}(g)), trail degradation, or garbage dumping. Deforestation through logging alone is a global issue, with concerning recent developments in the Amazonas~\cite{Deforestation_Fronts}, as well as locally in Romania. In addition to monitoring against these threats, tourists, cyclists and runners should be warned in advance about possible dangers present on the forest roads, including the detection and warning against common dangerous wildlife, such as bears or wild boars (Figure~\ref{fig:forestry_roads}(h)). Due to the large number and areal distribution of the roads, a monitoring approach based solely on human reconnaissance would be unscalable, making it impossible to detect, track and evaluate their condition in real-time.
 %Fixed monitoring stations equipped with cameras and measurement sensorics would be required in a large number, with permanent on-site maintenance, while also altering the natural state of the environment. 
 In order to overcome these challenges, we propose RovisLab AMTU (Autonomous Mobile Test Unit), which is a mobile robotic system project designed to monitor and map the state of forestry roads.
 
 %into a world-wide accessible Dynamic Forest Roads Map, accessible to individuals over the internet web-page. Evaluating in real-time the state of these large number of roads and acting accordingly would allow for a better protection of the Romanian ecosystem, allowing to better fight climate change and reversing the country’s carbon footprint.

\section{Related Work}
\label{sec:realated_work}
 
Unlike agricultural robots, robotic forest applications are less common. Such applications are environmental restoration, or wildfire fire-fighting. Hellstrom et al.\cite{hellstrom2009autonomous} explored the extent of automation and autonomy for forest vehicles, concentrating on technical principles that can be extended to autonomous forestry robotics.

%The authors of~\cite{tang2015slam} investigated the efficiency of SLAM-aided stem mapping for forest inventory. In a related study,~\cite{tremblay2020automatic} examined several methods of automatic 3D mapping for diameter measurements in inventory operations. The researchers used a Husky A200 mobile robot with a Velodyne LiDAR, IMU, wheel encoders, and an RGB mono monitor that skid steers.

In the embedded vision context, our task is similar to panoptic segmentation, first introduced in~\cite{Kirillov19} and which refers to the method of unifying two computer vision tasks, previously done separately. UPSNet \cite{xiong19upsnet} uses a single residual network as backbone, with two network heads for semantic and instance segmentation, respectively. In \cite{chen2020} the scene segmentation task has been improved by using position sensitive embeddings for instance segmentation, taking into consideration both the object appearance and its spatial location.

In the light of the advances mentioned above, we proposed a novel approach to embedded vision for self-driving off-road robots, which combines a learnable deep network with handcrafted features.

\section{Vision Dynamics System}
\label{sec:methodology}

The RovisLab AMTU robot from Figure~\ref{fig:rovislab_amtu} is a differential drive skid-steer wheeled mobile platform equipped with a $360^\circ$, $40$-channel Hessai Pandar Lidar, 4x e-CAM130A cameras and a Tinkerforge Inertial Measurement Unit (IMU). The control system is embedded on an NVIDIA AGX Xavier computer board, where resources are equally allocated for vision dynamics and low-level motion planning and control.

\begin{figure*}
	\centering
	\begin{center}
		\includegraphics[scale=0.9]{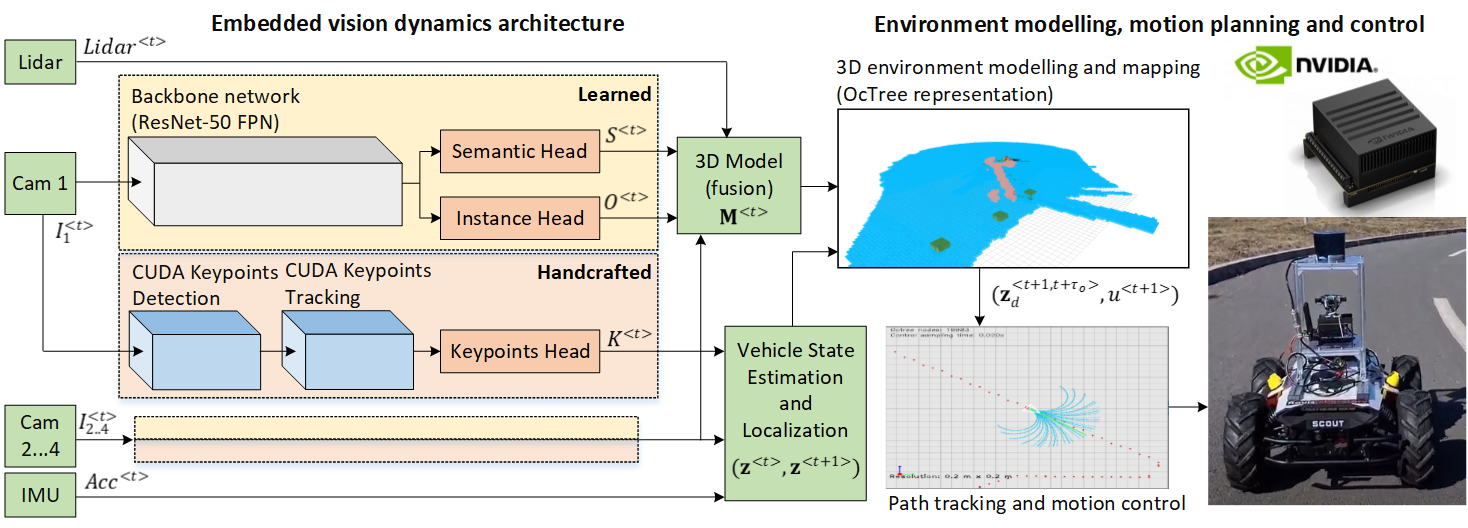}
		\caption{Embedded vision architecture for trajectory planning and motion control.}
        \label{fig:neural_network_diagram}
	\end{center}
    \vspace{-1em}
\end{figure*}

In order to achieve real-time capabilities, we propose the visual control architecture from Figure~\ref{fig:neural_network_diagram}. The motion of the robot is planned based on a give trajectory route and a real-time 3D reconstructed model of the surrounding environment. The trajectory is calculated within a 2D birds-eye view reprojected occupancy grid, where each cell encodes the occupancy of that location. We use an optimized version of the Dynamic Window Approach (DWA)~\cite{Fox_Dynamic_Window_Approach_1997} for generating possible driving trajectories. The best candidate trajectory is selected and executed via a Constrained Nonlinear Model Predictive Controller (NMPC)~\cite{ostafew-ijrr16}.

The robotic motion is directly dependent on the quality of the reconstructed 3D model, which in turn is calculated using our vision dynamics approach to real-time embedded vision. In the following, we describe our approach to address RovisLab's AMTU vision dynamics system design and training. 

\subsection{Embedded Vision Architecture}

We tackle RovisLab AMTU's problem of real-time robotic perception using the \textit{Vision Dynamics} theoretical framework~\cite{GRIGORESCU_Vision_Dynamics_2021}. Vision Dynamics is our concept for bridging the gap between perception systems and control algorithms. In our view, these two tasks are directly interlinked and should be optimized together by balancing safety, accuracy and real-time computation constraints.

Given a global reference trajectory $\vec{z}^{<t-\infty, t+\infty>}_{ref}$ and a set of observations comprising of camera images $I_{1,..,4}^{<t>}$, Lidar point clouds $Lidar^{<t>}$, accelerations $Acc^{<t>}$ obtained from the IMU sensor and the current GPS coordinates, the tasks are to:

\begin{enumerate}
    \item reconstruct the surrounding 3D environment $\vec{M}^{<t>}$, 
    \item localize and calculate the current state of the robot $\vec{z}^{<t>}$ on the global map, while predicting its next state $\vec{z}^{<t+1>}$ and 
    \item derive a safe, collision free state trajectory $\vec{z}^{<t+1, t+\tau_o>}$ which will be tracked by the motion controller using control signals $\vec{u}^{<t+1>}$.
\end{enumerate}

As shown in Figure~\ref{fig:rovislab_amtu}, each camera stream is processed using a dual vision dynamics architecture, comprised of a learnable multi-task Deep Neural Network (DNN) for semantic scene understanding and a handcrafted keypoints detector and tracker for Simultaneous Localization and Mapping (SLAM). We use four cameras rotated $90^\circ$ between each other. This allows us to get more keypoints that just using a single camera, thus increasing the precision of the SLAM component, while also providing a surround view for detecting deforestation.

The backbone of the DNN is a ResNet-50 with a Feature Pyramid Network (FPN). The two segmentation heads are used to predict at sampling time $t$ semantic $S^{<t>}$ and object instance IDs $O^{<t>}$ for each pixel $(x, y)$ in the input image:

\begin{equation}
    S^{<t>}(x, y) = c \text{ }\text{ }\text{ and }\text{ }\text{ } O^{<t>}(x, y) = k,
\end{equation}

\noindent where $c \in {1, ..., N}$ is the semantic class ID and $k \in \mathbb N$ is the instance ID.

We use the instance segmentation output to calculate 2D bounding boxes for each region of connected pixels:

\begin{equation}
    B^{<t>}(x, y) = (x_1, x_2, y_1, y_2, c),
\end{equation}

\noindent where $B^{<t>}(x, y)$ are the bounding boxes of the objects' instances and $(x_1, y_1)$ and $(x_2, y_2)$ are the coordinates of the top-left and bottom-right corners of a bounding box. The loss function is designed by dividing it into two sub-terms, one for each head of the DNN:

\begin{equation}
    L = \alpha L_{S} + \beta L_{O},
\end{equation}

\noindent where $L_{S}$ and $L_O$ are the semantic and instance softmax cross entropy losses, respectively, and $\alpha$ and $\beta$ are the corresponding weight factors.

The handcrafted pipeline is used for computing FAST keypoint features which are tracked in the input video streams using the pyramidal approximation of the Lucas-Kanade feature tracker~\cite{Nagy2020}. In this case, no training is required, since FAST is used to compute the probable corner points. The pose of the robot is computed using Perpective-N-Point mapping between the tracked keypoints in consecutive image frames and fusion with the acceleration data $Acc^{<t>}$ from the IMU. Due to its CUDA implementation, we have managed to track keypoints at a rate of $460$ FPS on all four input images combined.

Examples of semantic and instance segmentation results, as well as keypoints tracking are shown in Figure~\ref{fig:rovislab_amtu}(b).  The optimal balance between accuracy and real-time computation has been achieved both for segmentation and keypoints tracking for input images of size of $320 \times 240$ pixels.

The 3D model $\vec{M}^{<t>}$ is reconstructed using a-priori environment information. Namely, we project onto the 3D space the ground information based on the extrinsic parameters of the camera. Additionally, the 3D pose of the obstacles (e.g. pedestrians) are mapped as 3D bounding boxes using the 2D instances $O^{<t>}$ and the projected Lidar points into the corresponding camera images. For visualization purposes, the color code from Figure~\ref{fig:rovislab_amtu}(b) are the same as the ones used when rendering the 3D model. This is mostly visible in Figure~\ref{fig:neural_network_diagram}, where the inner 3D model shows the ground segmentation class in blue. 

\section{Experiments}
\label{sec:experiments}

The experiments have been performed in outdoor conditions around our research facility. The proposed system was benchmarked based on the accuracy of the embedded vision results, as well as with respect to the obtained computation time, all given in Table~\ref{tab:quant_results}. Since the handcrafted backbone has high computation speed, of around $2ms$, we have left it out of the evaluation table for clarity.

\begin{table}[]
\begin{tabular}{ccccc}
\hline
\multirow{2}{*}{Backbone} & \multirow{2}{*}{\begin{tabular}[c]{@{}c@{}}OA\\ (\%)\end{tabular}} & \multirow{2}{*}{\begin{tabular}[c]{@{}c@{}}mIoU\\ (\%)\end{tabular}} & \multirow{2}{*}{AP} & \multirow{2}{*}{\begin{tabular}[c]{@{}c@{}}Inference\\ Time(ms)\end{tabular}} \\
                          &                                                                    &                                                                      &                     &                                                                               \\ \hline
\multicolumn{5}{c}{\textbf{Full Image Size: 640 x 480 pixels}}                                                                                                                                                                                                              \\ \hline
ResNet-18 FPN             & 78.87                                                              & 74.72                                                                & 24.24               & 75.2                                                                          \\
ResNet-34 FPN             & 82.23                                                              & 75.14                                                                & 25.89               & 88.4                                                                          \\
ResNet-50 FPN             & 85.33                                                              & 76.85                                                                & 29.37               & 98.1                                                                          \\ \hline
\multicolumn{5}{c}{\textbf{Half Image Size: 320 x 240 pixels}}                                                                                                                                                                                                              \\ \hline
ResNet-18 FPN             & 72.34                                                              & 69.27                                                                & 17.85               & 39.4                                                                          \\
Resnet-34 FPN             & 74.73                                                              & 69.73                                                                & 18.43               & 42.4                                                                          \\
ResNet-50 FPN             & 79.66                                                              & 71.21                                                                & 19.87               & 48.0                                                                          \\ \hline
\end{tabular}
	\caption{Results of the proposed embedded vision system.}
	\label{tab:quant_results}
    \vspace{-1em}
\end{table}

Videos showing the proposed robotic system in action, as well as the output of the embedded vision dynamics system are available through the links provided in the abstract.

We have trained the DNN in two stages using transfer learning. In the first stage, the DNN has been trained on the $37$ classes of the Mapillary driving dataset~\cite{MVD2017}, with the purpose to create an initial set of weights. These weights are used in the second stage of training, where we train on our own environment data samples, also mapped to $37$ semantic segmentation classes.

Although currently we require manual labeling for the semantic segmentation training data, we intend to use either contrastive or BYOL-style self-supervised learning to automatically infer training labels from the odometry information of the robot.

The accuracy of 3D mapping in forest environments is dependent on the $6$h autonomy given by AMTU's battery and the precision of the GPS receiver used to correct drifting. Given the robot's maximum velocity of $1.5$m/s, we are able to cover $~32$km before recharging. The minimum inference time required by our target applications to navigate is directly linked to the velocity of the vehicle. Computations below $100$ms are enough, since the velocities obtained are well below the ones encountered by vehicles driving on public roads.  

The advantage of our approach over satellite surveillance is that deforestation can be detected in its earliest stages, before being noticeable in low-resolution satellite images. Additionally, certain areas can be occluded when viewed using a satellite.

\section{Conclusions}
\label{sec:conclusions}

We have presented an embedded solutions for self-driving on off-road terrain, with the purpose of protecting and maintaining forest roads in Romania. In order to achieve real-time capabilities, we have developed an embedded vision architecture which computes in real-time both learnable as well as handcrafted features. On short-term, we intend to experiment with RovisLab AMTU outside of our research infrastructure and on different environmental conditions.

{
\small
\bibliographystyle{ieee_fullname}
\bibliography{egbib}
}

\end{document}